\documentclass[letterpaper, 10 pt, conference]{ieeeconf}  

\IEEEoverridecommandlockouts                              

\usepackage{graphics} 
\usepackage{epsfig} 
\usepackage{times} 
\usepackage{amsmath} 
\usepackage{amssymb}  
\usepackage{multirow}
\usepackage[table,xcdraw]{xcolor}
\usepackage{hyperref}
\usepackage{amsmath} 
\usepackage{color}
\usepackage{booktabs} 
\usepackage{hyperref}
\usepackage{url}
\usepackage{cite}
\usepackage{algorithm, float}
\usepackage{algpseudocode}

\usepackage{caption}
\usepackage{subcaption}

\usepackage{etoolbox}
\usepackage{tikz}
\usetikzlibrary{tikzmark}
\usetikzlibrary{calc}

\title{\LARGE \bf
DBaS-Log-MPPI: Efficient and Safe Trajectory Optimization via Barrier States}

\author{Fanxin Wang$^{1\dagger}$, Haolong Jiang$^{1\dagger}$\, Chuyuan Tao$^{2}$, Wenbin Wan$^{3}$ and Yikun Cheng$^{2*}$
\thanks{$^{1}$Fanxin Wang, Haolong Jiang are with School of Advanced TechnoLogy, Xi'an Jiaotong Liverpool University
        {\tt\small Fanxin.Wang/Haolong.Jiang22@xjtlu.edu.cn}}%
\thanks{$^{2}$Chuyuan Tao, Yikun Cheng are with Department of Mechanical Science and Engineering, University of Illiois at Urbana-Champaign, Illinois, United States
        {\tt\small chuyuan2/yikun2@illinois.edu}}%
\thanks{$^{3}$Wenbin Wan is with Department of Mechanical Engineering,
University of New Mexico, New Mexico, United States
        {\tt\small wwan@unm.edu}}%
\thanks{$\dagger$These authors contributed equally to this work}%
\thanks{$*$Correspondance: {\tt\small yikun2@illinois.edu}}%
}

\begin{document}

\maketitle
\thispagestyle{empty}
\pagestyle{empty}

\begin{abstract}
Optimizing trajectory costs for nonlinear control systems remains a significant challenge. Model Predictive Control (MPC), particularly sampling-based approaches such as the Model Predictive Path Integral (MPPI) method, has recently demonstrated considerable success by leveraging parallel computing to efficiently evaluate numerous trajectories. However, MPPI often struggles to balance safe navigation in constrained environments with effective exploration in open spaces, leading to infeasibility in cluttered conditions. To address these limitations, we propose DBaS-Log-MPPI, a novel algorithm that integrates Discrete Barrier States (DBaS) to ensure safety while enabling adaptive exploration with enhanced feasibility. Our method is efficiently validated through three simulation missions and one real-world experiment, involving a 2D quadrotor and a ground vehicle navigating through cluttered obstacles. We demonstrate that our algorithm surpasses both Vanilla MPPI and Log-MPPI, achieving higher success rates, lower tracking errors, and a conservative average speed.

\end{abstract}

\section{INTRODUCTION}

With the increasing focus on robotic control, designing safe and reliable control methods for autonomous robots operating in unknown environments—while maintaining real-time performance—remains a significant challenge \cite{dai2020rgb}. Successful navigation requires robots to accurately detect both static and dynamic obstacles, including convex and non-convex shapes, using appropriate sensors. Moreover, robots must dynamically re-plan their trajectories to avoid local optima, prevent collisions, and reach their target locations efficiently. These requirements give rise to a complex control optimization problem that is inherently difficult to solve in real-time \cite{claussmann2019review}.

Model predictive control (MPC) \cite{kouvaritakis2016model}, a well-established control framework, offers robust navigation capabilities by handling obstacles under both hard and soft system constraints. MPC employs a receding horizon strategy to generate a sequence of control inputs over a predefined prediction window. Only the first input is executed, while the remaining inputs serve as warm starts for subsequent optimization steps. MPC methods can be broadly categorized into gradient-based and sampling-based approaches. Gradient-based MPC leverages optimization techniques to generate smooth, collision-free trajectories while enforcing system constraints \cite{gaertner2021collision}. However, this approach relies on strong assumptions, such as the differentiability and convexity of cost functions, system constraints, and obstacle geometries—assumptions that often limit real-world applicability. Recent efforts to address nonconvexity include reformulating non-differentiable constraints into differentiable forms \cite{zhang2020optimization} and constructing convex hulls in cost maps \cite{brudigam2020grid}. However, these solutions remain computationally demanding.

\begin{figure}[t]
    \centering
    \includegraphics[width=0.9\linewidth]{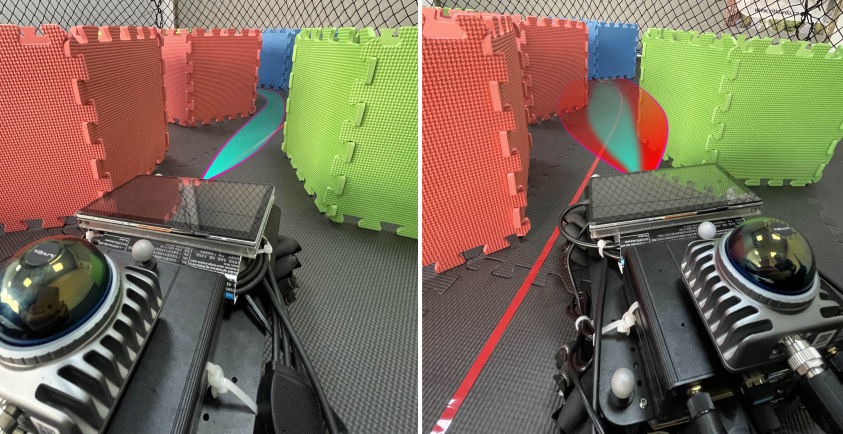} 
    \caption{Demonstration of trajectory sampling: ours (left) vs. vanilla MPPI (right).}
    \label{fig:fig1}
    \vspace{-8mm}
\end{figure}

In contrast, sampling-based MPC, such as the model predictive path integral (MPPI) \cite{williams2016aggressive}, circumvents gradient computations and efficiently handles nonconvex environments. By evaluating trajectories sampled from a probability distribution, MPPI explores diverse state-space possibilities to identify optimal paths. However, its performance is highly dependent on the sampling distribution and remains susceptible to local optima \cite{williams2018information}. Inspired by \cite{mohamed2022autonomous}, we propose an enhanced MPPI framework that employs a normal-log-normal mixture distribution to improve trajectory sampling efficiency and enhance adaptive exploration around local optima (\ref{fig:fig1}). Furthermore, we simplify safety constraints by integrating barrier states \cite{kuperman2023improved}, transforming the original optimization constraints into a state stabilization problem. This reformulation reduces optimization complexity, enabling our method to outperform MPPI-based approaches \cite{williams2016aggressive, mohamed2022autonomous} in both computational efficiency and navigation performance.

This paper is organized as follows: Section \ref{preliminaries} introduces the fundamentals of MPPI and barrier states. Section \ref{methodologies} presents the proposed DBaS-Log-MPPI method. In Section \ref{results}, we validate our approach in simulation with 2D quadrotor and ground vehicle, comparing its performance against Vanilla MPPI and log-MPPI. Furthermore, we demonstrate our method’s real-time performance in a real-world experiment on the ground vehicle "Antelope", showcasing the effectiveness in challenging trajectory tracking scenarios.


\section{Related Works}

Real-time robotic navigation in the presence of cluttered obstacles is challenging, necessitating rapid adaptation to both static and dynamic obstacles and the ability to reliably reach goals without becoming trapped in local optima. Current approaches to tackle this task primarily fall into two categories: control-oriented methods and learning-based approaches.

Control-oriented approaches, notably Model Predictive Control (MPC) \cite{kouvaritakis2016model}, address robotic navigation by solving constrained optimization problems online over a finite horizon. Prior to MPC, classical control methods such as the Linear Quadratic Regulator (LQR) \cite{scokaert1998constrained} provided optimal open-loop solutions for trajectory tracking tasks. However, these approaches are limited due to their offline computational nature and inability to handle dynamic obstacles and environmental uncertainties in real-time. Moreover, LQR inherently assumes linear system dynamics or requires linearization for nonlinear scenarios, making it unsuitable for complex, nonlinear, high-dimensional systems. Finite-horizon LQR or MPC extend these capabilities by solving constrained optimization problems online, explicitly managing state and input constraints. Gradient-based MPC methods necessitate differentiable constraints and cost functions, enabling smooth control sequences. However, these methods impose strong assumptions such as differentiability and convexity of cost functions, thus limiting their applicability in scenarios involving non-convex and non-differentiable objectives \cite{claussmann2019review}. Several approaches have been developed to overcome these limitations. For example, \cite{brudigam2020grid} introduced computationally efficient stochastic MPC using chance constraints that activate only when necessary. \cite{liu2017combined} proposed data partitioning from LiDAR sensors into convex shapes to enforce convexity. Sampling-based MPC methods, such as Model Predictive Path Integral (MPPI), mitigate these issues by employing stochastic sampling without strict differentiability and convexity requirements. Nonetheless, MPPI methods can produce non-smooth trajectories and potentially infeasible control sequences, particularly due to poor sampling distributions. Addressing these issues, \cite{mohamed2022autonomous} modified the sampling distribution strategically, while \cite{williams2018information} proposed methods to smooth control sequences.

Alternatively, learning-based methods, primarily represented by reinforcement learning (RL) \cite{sutton1998reinforcement}, offer promising directions to tackle these navigation challenges \cite{yen2004reinforcement}. RL utilizes offline data-driven optimization during training, circumventing computationally intensive online optimization, thus enhancing real-time performance compared to MPC \cite{song2023reaching}. Despite these advantages, training RL agents for effective obstacle avoidance remains challenging because obstacle avoidance objectives are typically represented as soft costs within reward functions \cite{yang2023safety}. This presents difficulty in enforcing strict safety constraints. Control Barrier Functions (CBFs) have emerged as robust safety mechanisms integrated with RL as safety filters, ensuring trajectories remain within predefined safe sets during training and deployment \cite{cheng2023safe, cheng2019end}. Nonetheless, RL methods encounter significant performance degradation due to environmental perturbations, particularly when policies trained in simulation are transferred to real-world robotic systems. Efforts to improve safety and robustness of RL through adaptive control were presented in \cite{yikunRL1-ieee}. Other methods such as domain randomization \cite{tobin2017domain} and meta-learning \cite{chen2021meta} have been utilized to enhance long-term robustness and safety but may lack ability to robustify RL policies instantly without the prior knowledge of the uncertainties.


\section{Preliminaries} \label{preliminaries}

In this section, we formalize the optimal control problem introduce the two primary methodologies utilized in this work: (i) trajectory optimization based on Model Predictive Path Integral (MPPI), and (ii) safety enforcement through Discrete Barrier States (DBaS).

\subsection{MPPI-Based Control with Constraints}

We consider a discrete-time nonlinear control system described by
\begin{equation} \label{eq:dynamics}
  x_{k+1} = f(x_k, u_k),
\end{equation}
where \( x_k \in \mathbb{R}^n \) denotes the state at time \( k \), and the control sequence with dimension $m$ over time horizon $N$ is given by 
\[
U = \{u_0, u_1, \dots, u_{N-1}\} \subset \mathbb{R}^{m \times N}.
\]
The resulting state trajectory is denoted by 
\[
X = \{x_0, x_1, \dots, x_{N-1}\}.
\]

The objective is to find a control sequence \( U \) that steers the system from an initial state \( x_c \) to a desired terminal state \( x_f \) while avoiding collisions and satisfying constraints. Unlike gradient-based Model Predictive Control (MPC) approaches, the Model Predictive Path Integral (MPPI) method avoids derivative computations, making it suitable for highly nonlinear, non-convex objectives.

The MPPI optimization problem is formulated as \cite{williams2017model}:
\begin{equation} \label{eq:mppi}
\begin{aligned}
  \min_{U} \quad & J = \mathbb{E}\Bigg[ \phi(x_N) + \sum_{k=0}^{N-1} \Bigg( q(x_k) + \frac{1}{2}\, v_k^T R\, v_k \Bigg) \Bigg] \\
  \text{subject to} \quad & x_{k+1} = f(x_k, v_k), \quad x_0 = x_c, \\
  & G(u_k) \leq 0,\quad h(x_k) \geq 0,\quad k = 0,1,\dots,N-1,
\end{aligned}
\end{equation}
where \( \phi(x_N) \) is the terminal cost, \( q(x_k) \) is a state-dependent running cost, \( R \) is a positive definite control weighting matrix, and \( v_k = u_k + \delta u_k \) with \( \delta u_k \sim \mathcal{N}(0, \Sigma_u) \) representing the injected noise.

At each time step, MPPI generates \( M \) trajectories in parallel by perturbing the control inputs. The cost-to-go for a sampled trajectory \( \tau \) is defined as
\begin{equation} \label{eq:cost_to_go}
  \tilde{S}(\tau) = \phi(x_N) + \sum_{k=0}^{N-1} \left( q(x_k) + \gamma\, u_k^T \Sigma_u^{-1} u_k \right),
\end{equation}
where \( \gamma \) governs the influence of the control cost. The optimal control update is then obtained via a weighted average \cite{williams2017model}:
\begin{equation} \label{eq:control_update}
  u_k^* = u_k + \frac{\sum_{m=1}^M \exp\Big(-\frac{1}{\lambda} \big(\tilde{S}(\tau_{k,m}) - \rho\big)\Big) \, \delta u_{k,m}}{\sum_{m=1}^M \exp\Big(-\frac{1}{\lambda} \big(\tilde{S}(\tau_{k,m}) - \rho\big)\Big)},
\end{equation}
where \( \rho = \min_m \tilde{S}(\tau_{k,m}) \) and \( \lambda > 0 \) is the inverse temperature parameter. Finally, a smoothing operation (e.g., Savitzky--Golay filtering) is applied to reduce noise, and the control input \( u_0^* \) is applied in a receding horizon framework.

\subsection{Discrete Barrier States for Safety}

Safety is ensured if the state remains within a safe set \( S \subset \mathbb{R}^n \) for all time steps . Traditional Control Barrier Functions (CBFs) enforce safety through solving complex optimization problems\cite{ames2019control}, while barrier states provide a more intuitive and direct representation of safety constraints by explicitly encoding the collision assessment into states, which can be more flexible and computational effective in handling dynamic environments \cite{khan2020gaussian}.

Our approach embeds safety constraints into the system dynamics via DBaS. We define a barrier function \( B: S \to \mathbb{R} \) that is smooth and strongly convex on the interior of \( S \) and diverges as the state approaches the boundary \( \partial S \). Composing \( B \) with a function \( h(x) \) that characterizes the safety condition, we define the barrier state as
\begin{equation} \label{eq:barrier_state}
  \beta(x_k) = B\big(h(x_k)\big).
\end{equation}
The boundedness of \( \beta(x_k) \) ensures \( h(x_k) \geq 0 \).

To integrate safety, we augment the state with the barrier state to form an extended state \( \hat{x}_k = \begin{bmatrix} x_k \\ \beta(x_k) \end{bmatrix} \). The safety-embedded system dynamics are then given by
\begin{equation} \label{eq:augmented_dynamics}
  \hat{x}_{k+1} = \hat{f}(\hat{x}_k, u_k).
\end{equation}

The barrier state is updated as\cite{almubarak2023barrier}
\begin{equation} \label{eq:barrier_update}
  \beta(x_{k+1}) = B\big(h(f(x_k, u_k))\big) - \gamma\Big(B\big(h(x_d)\big) - \beta(x_k)\Big),
\end{equation}
where \( \gamma \in (0,1) \) is a tunable parameter and \( x_d \) denotes the desired equilibrium. For multiple constraints, a fused barrier state is constructed and appended to the state. The system is safe if and only if \( \beta(x_k) < \infty \) for all \( k \).

\begin{figure*}[ht]
  \centering
  \vspace{2mm}
  \includegraphics[width=0.85\hsize]{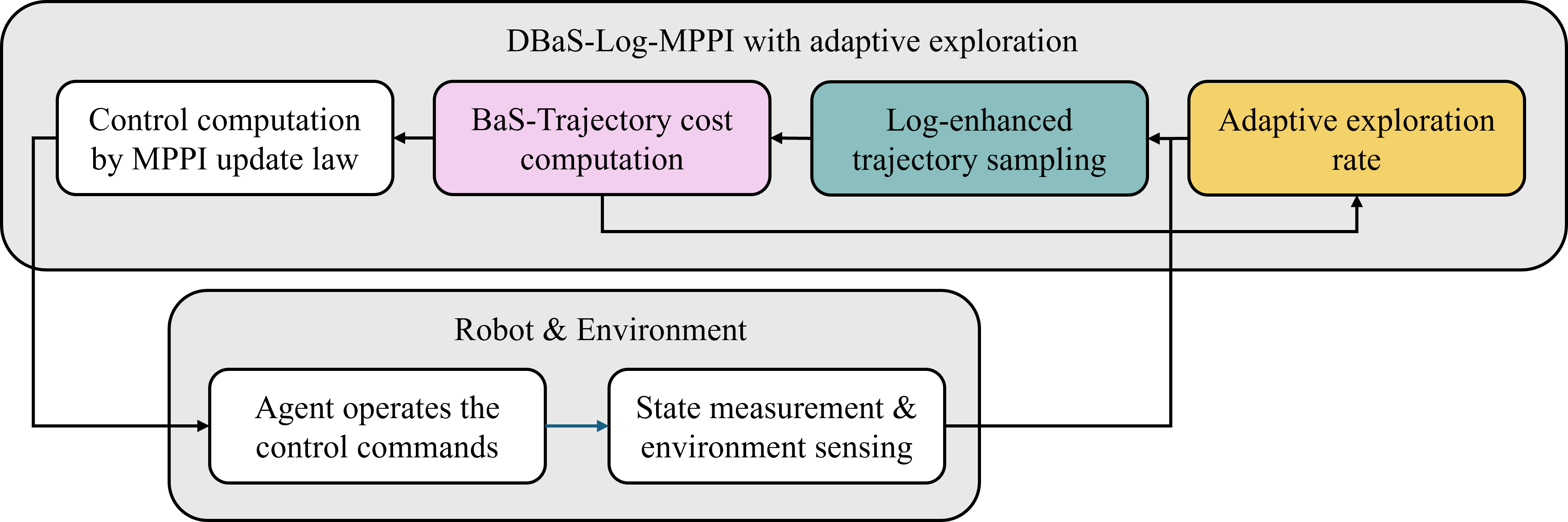}
  \caption{Proposed DBaS-Log-MPPI control scheme.}
  \label{fig2}
  \vspace{-8mm}
\end{figure*}

\section{DBaS-Log-MPPI} \label{methodologies}
In this section, we will introduce the detailed algorithm of DBaS-Log-MPPI, the block diagram could be referred to Fig.~\ref{fig2}. Our DBaS-Log-MPPI controller introduces three key enhancements:

\textbf{Integrating of Barrier States:} By embedding barrier states into the system dynamics, safety concerns are directly addressed as part of the controlled states. This allows the controller to maintain safety as the cost of these barrier states is incorporated into the running cost, replacing the traditional collision impulse-indicator cost. This ensures that safety is inherently and comprehensively considered in the optimization process.

\textbf{Adaptive Exploration:} The cost mechanism guides our adaptive exploration strategy. As the trajectories closer to obstacles incur higher running costs, reflecting increased collision risk. To circumvent local optima—often encountered without finely tuned injected noise variance—the algorithm automatically increases the exploration rate from the dynamic running cost, encouraging diverse trajectory sampling and facilitating the discovery of approximately optimal trajectories \cite{yin2022trajectory}. Once the initial broad search identifies promising regions, the exploration rate is reduced in subsequent steps, enabling precise convergence to near-optimal solutions while preserving safety guarantees.

\textbf{Enhanced Feasibility with Log-sampling Strategy:} While adaptive exploration with a larger covariance matrix $\Sigma_u$ can lead to input chattering and potential violations of system constraints, particularly in cluttered and input-constrained environments, the Log-sampling strategy\cite{mohamed2022autonomous} offers a significant improvement. This strategy employs a mixture of normal and Log-normal distributions ($\mathcal{NLN}$ mixture) for trajectory sampling, ensuring more efficient state-space exploration with lower variance. By respecting system constraints and reducing the risk of local minima, the Log-sampling strategy enhances feasibility and achieves superior performance with fewer samples.

\subsection{DBaS augmentation with the running cost}
In the Vanilla MPPI framework, constraints are typically evaluated using weighted indicator terms. These terms impose an impulse-like penalty upon constraint violation, effectively acting as a hybrid approach between hard and soft constraints. However, this method has a critical limitation: it provides no risk information until a constraint is actually violated\cite{yin2023shield}. This poses significant challenges for trajectory optimization, particularly in scenarios involving narrow passages or high speeds. Specifically, until a collision is predicted to occur, all trajectories are evaluated solely based on state-dependent costs, without any consideration of proximity to constraints or potential risks. This lack of proactive safety assessment can lead to unsafe trajectories in complex environments, as illustrated in Fig.~\ref{fig:coll} in the results section. To address this limitation, we propose the following Discrete Barrier State augmentation in the running cost.

Starting at each loop, the control input perturbations $\delta u_k$ and their corresponding trajectories are sampled according to the desired exploration rate (introduced in Secton\ref{adaptive exploration}), guided by the DBaS augmented system dynamics. The DBaS ensures that all sampled trajectories remain safe by embedding safety constraints directly into the system dynamics. This allows the controller to evaluate proximity to obstacles and collision risk continuously, rather than relying on impulse-like penalties upon constraint violation:

\begin{equation}\label{eq:MPPI_BaS}
    \begin{aligned}
        &\hat{x}_{k+1} = \hat{f}(\hat{x}_k, v_k), \\
        &v_k = u_k + \delta u_k,
    \end{aligned}
\end{equation}
where $\hat{x}_k$ represents the augmented system state embedded with the DBaS. 
The stochastic sampling of candidate trajectories and their subsequent cost evaluations remain unchanged, ensuring that the convergence behavior of the Model Predictive Path Integral (MPPI) algorithm is preserved \cite{williams2018information}. The integration of DBaS does not compromise the asymptotic convergence properties of MPPI, as it seamlessly incorporates safety considerations into the cost structure. This approach guarantees strict constraint satisfaction while maintaining the stability, robustness, and convergence guarantees inherent to the underlying optimization framework.

Additionally, DBaS provides a significant advantage in controlling high relative-degree systems\cite{almubarak2022safety}, where constraints at the output level are influenced through multiple derivatives of the control input. Traditional methods often require defining barrier functions at higher derivatives and incorporating carefully chosen exponential terms to achieve exponential stability of constraint enforcement. In contrast, DBaS leverages costate dynamics to naturally encapsulate the interdependencies between states and constraints.

By sampling the barrier state alongside the perturbed input at each time step, we ensure that safety constraints are satisfied throughout the trajectory without the computational burden of solving the barrier state equation at every iteration. Under this augmented dynamics, $M$ number of samples are generated, and the cost-to-go of the state-dependent cost for each trajectory is calculated as follows:
\begin{equation}\label{eq:cost-to-go_bas}
    \Tilde{S}(\tau) = C_B(\hat{X})+\phi(\hat{x}_N) + \sum_{k=0}^{N-1} q({x}_k)+\gamma u_{k}^T \Sigma_u^{-1} v_k,
\end{equation}
where the term $C_B(\hat{X}) = \sum_{k=0}^{N} R_B \space w({x}_k)$ represents the barrier state cost along the trajectory. $R_B$ is the cost weight on the barrier state. The barrier state itself is inherently positive, so there is no need to square it to ensure positive definiteness in the cost term. The cost-to-go function implies that collisions are penalized in the sampled trajectory, even though only the first control instance is applied to the system (as described in \eqref{eq:MPPI_BaS}). 

With equation \eqref{eq:control_update}, the "near-optimal" controls sequence $\mathbb{U^*} = (u_0^*, u_1^*, \ldots, u^*_{N-1}) \in \mathbb{R}^{m \times N}$ is calculated and then applied to the system dynamic equation \eqref{eq:MPPI_BaS} with the cost $J= \mathbb{E}[\Tilde{S}(\tau)]$ being optimized.

\subsection{Adaptive trajectory sampling}\label{adaptive exploration}
$M$ trajectories are sampled using the same covariance matrix $\Sigma_u$
to derive "near-optimal" controls in the Vanilla MPPI. This approach takes the advantage of high computational efficiency by avoiding the need for explicit gradient calculations, as required in traditional optimization methods. However, applying a fixed exploration rate determined by the same covariance matrix $\Sigma_u$ can lead to two significant issues in tightly-constrained system-environment scenarios:

Using a relatively small $\Sigma_u$ facilitates fine trajectory evaluation in loosely-constrained tasks, as most sampled trajectories are closely clustered. However, in tightly-constrained scenarios, this limited exploration can result in entrapment in local optima. The controller fails to explore sufficient portions of the state space, making it difficult to identify feasible paths through narrow gaps or complex obstacles. A detailed visual explanation could be found in Fig.~\ref{fig:sleep} in the results section.

Using a relatively large $\Sigma_u$ increases the likelihood of finding a feasible path in tightly-constrained scenarios, as the sampled trajectories explore a broader hyper-space of the controller. However, this approach can underperform in open spaces, where widely dispersed trajectories lead to coarse optimization results rather than precise path-following with accurate tracking. Additionally, excessive exploration can introduce input chattering and increase the risk of violating system constraints. A detailed visual explanation could be found in Fig.~\ref{fig:long} in the results section.

We propose an adaptive trajectory sampling framework leveraging the DBaS property to ensure constraint satisfaction. By embedding DBaS into the nominal state, we ensure constraints are inherently satisfied, enabling flexible adjustment of $\Sigma_u$ in tightly constrained environments. As the system nears forbidden regions, the barrier function $h(x_k)$ decreases, causing $B\circ h(x_k)$ to grow significantly, thereby increasing $w(\hat{x}_k)$. To balance exploration and precision, we introduce an adaptive exploration rate $S_e$, defining the injected disturbance $\delta \hat{u}_k$ as a $\mathcal{NLN}(0, S_e\Sigma_u)$ distribution (introduced in Section\ref{Log-sampling}). 

\begin{equation}\label{eq:scaling factor}
    S_e = \mu \space Log(e+ C_B(\hat{X}^*)),
\end{equation}
where $\mu \in (0,1)$ is the coarseness factor, which determines the accuracy of trajectory tracking in free space. To prevent overly aggressive exploration, a Logarithmic function is applied to $S_e$. This is particularly important in tightly-constrained scenarios, where the barrier cost increases rapidly as the system approaches obstacles or limits. The Logarithmic transformation ensures smooth and controlled exploration behavior.

After each optimization step, the barrier cost $ C_B(\hat{X}^*)$ is evaluated for the trajectory using the current optimal control input $\mathbb{U}^*$, dynamically adjusting $S_e$ to maintain a balance between exploration and constraint satisfaction.

\begin{algorithm}[!hb]
\caption{DBaS-Log-MPPI Algorithm}
\label{alg:the_alg}
\begin{algorithmic}[1]
\Require $f$: Transition Model; \\
$M$: Number of samples; \\
$T$: Number of time-steps; \\
$(u_0, u_1, \ldots, u_{T-1})$: Initial control sequence; \\
$\mu, \Sigma_{u}, \phi, q, \gamma, R_B$: Cost functions/parameters; \\
SGF: Savitsky-Golay convolutional filter;
\While{task not completed}
    \State $x_0 \gets \text{GetStateEstimate}()$
    \For{$k \gets 0$ to $M-1$}
        \State $x \gets x_0$
        \State Sample  $\delta u_t^k \sim \mathcal{NLN}(0, S_e\Sigma_u)$
        \For{$t \gets 1$ to $T$}
            \State $v = u+\delta u$
            \State $x \gets f(x, v)$
            \State $w \gets \sum\mathbf{B} \circ h^{(i)} \circ  f(x_t, v_t)-\gamma(\beta(x_d)-\beta(x_t))$
            \vspace{-2mm}
            \State $\hat{x} = \begin{bmatrix} x \\ w \end{bmatrix}$
            \State $\Tilde{S}_k += q(\hat{x})+\gamma u_{}^T S_e^{-1}\Sigma_u^{-1} v+ R_B \space w({x}_k)$ 
        \EndFor
        \State $\Tilde S_k += \phi(x)$
    \EndFor
    \For{$t \gets 0$ to $T-1$}
        \State $U^* \gets U + \text{SGF}  \frac{\sum_{m=1}^{M}\exp(-(1/\lambda)(\Tilde{S}_k-\rho))\delta u_{k,m}}{\sum_{m=1}^{M}\exp(-(1/\lambda)(\Tilde{S}_k-\rho))}$
    \EndFor
        \State SendToActuators($u_0$)
    \For{$t \gets 1$ to $T$}
        \State $\hat{x}= \hat{f}(\hat{x}, u*) $
        \State $C_B(\hat{X}^*)+=R_B \space w({x}_k)$
    \EndFor
    \State $S_e = \mu \space log(e+ C_B(\hat{X}^*))$
    \For{$t \gets 1$ to $T-1$}
            \State $u_{t-1} \gets u_t$
        \State $u_{T-1} \gets \text{Initialize}(u_{T-1})$
    \EndFor
\EndWhile
\end{algorithmic}
\end{algorithm}

\subsection{Enhanced feasibility with Log-sampling strategy}\label{Log-sampling}
Augmenting DBaS with adaptive trajectory sampling \cite{wang2025mppi} has demonstrated strong performance in obstacle avoidance tasks. However, systems operating at high speeds may still face "emergency avoidance" scenarios when fast-approaching obstacles, due to the limited prediction horizon $N$. Infeasibility arises when the explored trajectories cannot turn or brake quickly enough within the constrained sampling space. The Log-sampling strategy\cite{mohamed2022autonomous} addresses this limitation by employing a mixture of normal and Log-normal distributions for trajectory sampling. This approach enhances exploration efficiency, enabling faster and more feasible responses to cluttered obstacles, particularly in high-speed scenarios. By combining DBaS with Log-sampling, the framework achieves improved robustness and feasibility in complex, time-critical environments.

Suppose that \( X_1 \sim  \mathcal{N}(\mu_{n}, \sigma_{n}) \)  (normal distribution) and \( X_2 \sim  \mathcal{LN}(\mu_{ln}, \sigma_{ln}) \) (Log-normal distribution), then the random variable \( Z=X_1X_2 \sim  \mathcal{NLN}(\mu_{nln}, \sigma_{nln}) \) is defined as the normal-log-normal ($\mathcal{NLN}$) distribution. The corresponding mean $\mu_{nln}$ and variance $\sigma_{nln}$ are \cite{yang2008normal}:

\begin{equation}\label{eq:nln mean and var}
    \begin{aligned}
        &\mu_{nln} = \mathbb{E}(X_1X_2) = \mathbb{E}(X_1)\mathbb{E}(X_2)=\mu_{n}e^{\mu_{ln}+0.5\sigma_{ln}^2}, \\
        &\begin{aligned}
            \sigma_{nln}&=\mathrm{Var}(X_1X_2)=\mathbb{E}(X_1^2)\mathbb{E}(X_2^2)-\mathbb[{E}(X_1X_2)]^2\\
            &=(\mu_{n}^2+\sigma_{n}^2)e^{2\mu_{ln}+2\sigma_{ln}^2}-\mu_{n}^2e^{2\mu_{ln}+\sigma_{ln}^2}.
        \end{aligned}
    \end{aligned}
\end{equation}

The detailed injected disturbance $\delta \hat{u}_k$ is sampled under this distribution $\delta \hat{u}_k \sim \mathcal{NLN}(\mu_{nln}, \sigma_{nln})$. It is pointed out that even if $\mu_n = 0,$ $\mu_{nln} =0$, indicating $\delta \hat{u}_k$ would be a symmetric distribution around 0. Even with a same value of $\mu$, normal distribution would explore less action space and state space compared to normal-log-normal distribution\cite{mohamed2022autonomous}.
\subsection{MPPI-DBaS with adaptive exploration algorithm}
Consistent with the approach outlined in Fig.~\ref{fig2}, we present the detailed DBaS-Log-MPPI Algorithm\ref{alg:the_alg} as above.

\section{Experimental Results} \label{results}

In this section, we design three distinct navigation missions in simulation to evaluate the proposed algorithm's performance for a quadrotor and a ground vehicle. For each mission, 100 experiments are conducted, and the results are summarized in Table~\ref{tab:mission_performance_comparison}. Key metrics include task success rate (collision-free), relative path tracking error, and average tracking speed.

Additionally, the algorithm is implemented on the "Antelope" ground vehicle in a real-world scenario. The experiments demonstrate the algorithm's effectiveness in performing safe trajectory optimization in challenging navigation tasks.

\begin{table}[ht]
\renewcommand{\arraystretch}{1.2}
\centering
\caption{Simulation Mission Performance Comparison}
\label{tab:mission_performance_comparison}
\begin{tabular}{|c|c|c|c|}
\hline
\multirow{2}{*}{\textbf{\textit{Indicator}}} & \multicolumn{2}{c|}{\textbf{\textit{Baselines}}} & \textbf{\textit{Ours}} \\ \cline{2-4}
                            & \textit{Vanilla MPPI} & \textit{Log-MPPI} & \textit{DBaS-Log-MPPI} \\ \hline
\multicolumn{4}{|c|}{\textbf{Mission 1:}  Quadrotor , $v_{set} = 1$ m/s} \\  \hline
   \textbf{\textit{Success}} [\%]   & 0  & 43  & \cellcolor{gray!15} 100 \\ 
   \textbf{\textit{Error}} [m]  & -  & 5.360  & \cellcolor{gray!15} 2.889 \\ 
   \textbf{\textit{Avg vel} } [m/s]  &  -  & \cellcolor{gray!15} 0.864  &  0.721 \\ \hline
\multicolumn{4}{|c|}{\textbf{Mission 2:}  Ground vehicle , $v_{set} = 5$ m/s} \\ \hline
   \textbf{\textit{Success}} [\%]   & 27  & 74 & \cellcolor{gray!15} 100 \\ 
   \textbf{\textit{Error}} [m]  & 0.814 & 0.741 & \cellcolor{gray!15} 0.703 \\ 
   \textbf{\textit{Avg vel}} [m/s]   & \cellcolor{gray!15} 4.781  & 4.679 & 4.690 \\ \hline
\multicolumn{4}{|c|}{\textbf{Mission 3:}  Ground vehicle , $v_{set} = 8$ m/s} \\ \hline
   \textbf{\textit{Success}} [\%]  & 0 & 35  & \cellcolor{gray!15} 70  \\ 
   \textbf{\textit{Error}} [m]  & -  & \cellcolor{gray!15} 0.240  &  0.341 \\ 
   \textbf{\textit{Avg vel}} [m/s]  & -  & 7.347  & \cellcolor{gray!15} 7.504  \\ \hline
\end{tabular}

\end{table}

\subsection{Quadrotor navigation simulation}

We first design a simulation scenario for the 2D quadrotor experiment. We consider the nonlinear dynamic model in the following format:

\begin{equation}\label{veh_dyn}
    \begin{bmatrix}
x_{k+1}\\ 
y_{k+1}\\ 
\theta_{k+1}\\
v_{x_{k+1}}\\
v_{z_{k+1}}

\end{bmatrix} =   \begin{bmatrix}
x_{k} + v_x\Delta t\\ 
y_{k} + v_y\Delta t\\ 
\theta_{k} + \omega\Delta t\\
v_{x_{k}} - \frac{mg+u_{t}}{m}sin(\theta_{k})\Delta t\\
v_{z_{k}} + (\frac{mg+u_{t}}{m}cos(\theta_{k})-g)\Delta t
\end{bmatrix}
\end{equation}

where $\mathbf{x_k}=[x,z,\theta,v_x,v_z]$ represents the position $x$, position $z$, pitch $\theta$, velocity $v_x$ and velocity $v_z$ expressed in the world frame. The control $\mathbf{u_k}=[\omega, u_t]$ represents the control signal of pitch rate and throttle (thrust force over gravity) with limits of $[\pm 4rad/s, \pm 0.981N]$. The quadrotor are set with 0.4$m$ frame length, 0.5$kg$ mass, 0.005$kgm^2$ moment of inertia. The gravity $g$ is set to 9.81$m/s^2$. We propose a specific quadrotor model where collision detection is facilitated by defining seven shape points. These shape points are positioned at both propellers and each quarter points along the frame. The obstacle function is defined as:
\begin{equation}
    \begin{aligned}
   h(x_i) =\| x_i - x_c^j \|_2^2 - r_c^j{^2} \geq 0, \quad \text{for } &i \in \{ 1, \dots \, 7\}, \\
   &j\in \{1, \dots \}
   \end{aligned}
\end{equation}
where $x_i$ denotes the shape points and $x_c^i$ is the center of $i$th circle obstacle with radius of $r_c$.

Sampling covariance matrix $\Sigma_u$ are set as $\begin{bmatrix} 0.4& 0 \\ 0& 0.12 \end{bmatrix}$, $\gamma$ are set to be 2 for all three algorithms. Coarseness factor $\mu$ is set to be 0.4 in DBaS-Log-MPPI. The real-time execution is carried out on an NVIDIA GeForce GTX 3070 laptop GPU, where all algorithms were written in Python. The prediction time horizon $N$ is set to 20 steps, the sampling time $\Delta t$ is set to be 0.02$s$ and parallel sampled trajectories number $M$ is set to 1024. The average processing time is 0.017 s, with no significant differences observed among the three algorithms.

\begin{figure}[!hb]
\centering
\begin{subfigure}{0.235\textwidth}
    \centering
    \includegraphics[width=\textwidth]{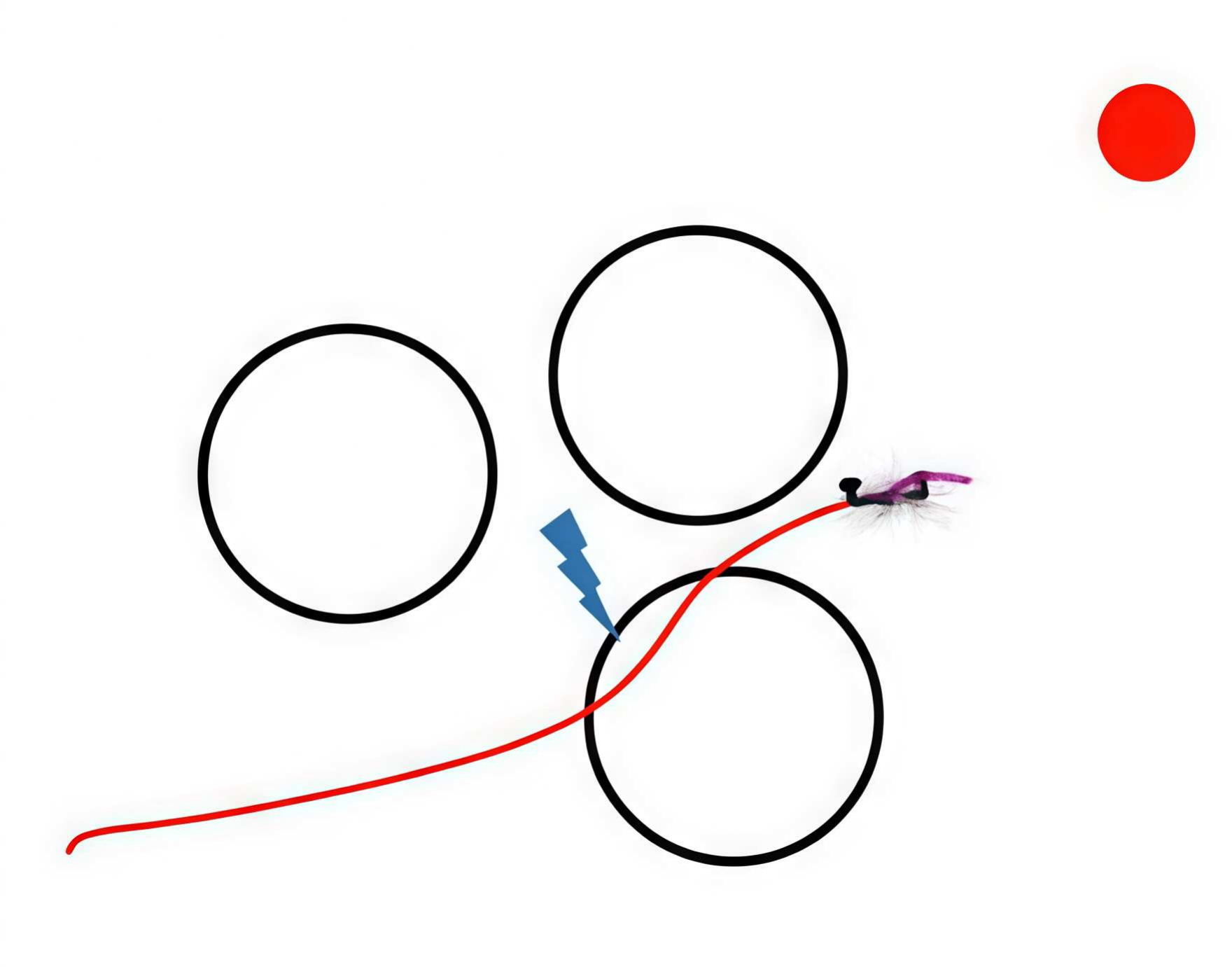} 
    \caption{Vanilla MPPI collision.}
    \label{fig:coll}
\end{subfigure} \hfill
\begin{subfigure}{0.235\textwidth}
    \centering
    \includegraphics[width=\textwidth]{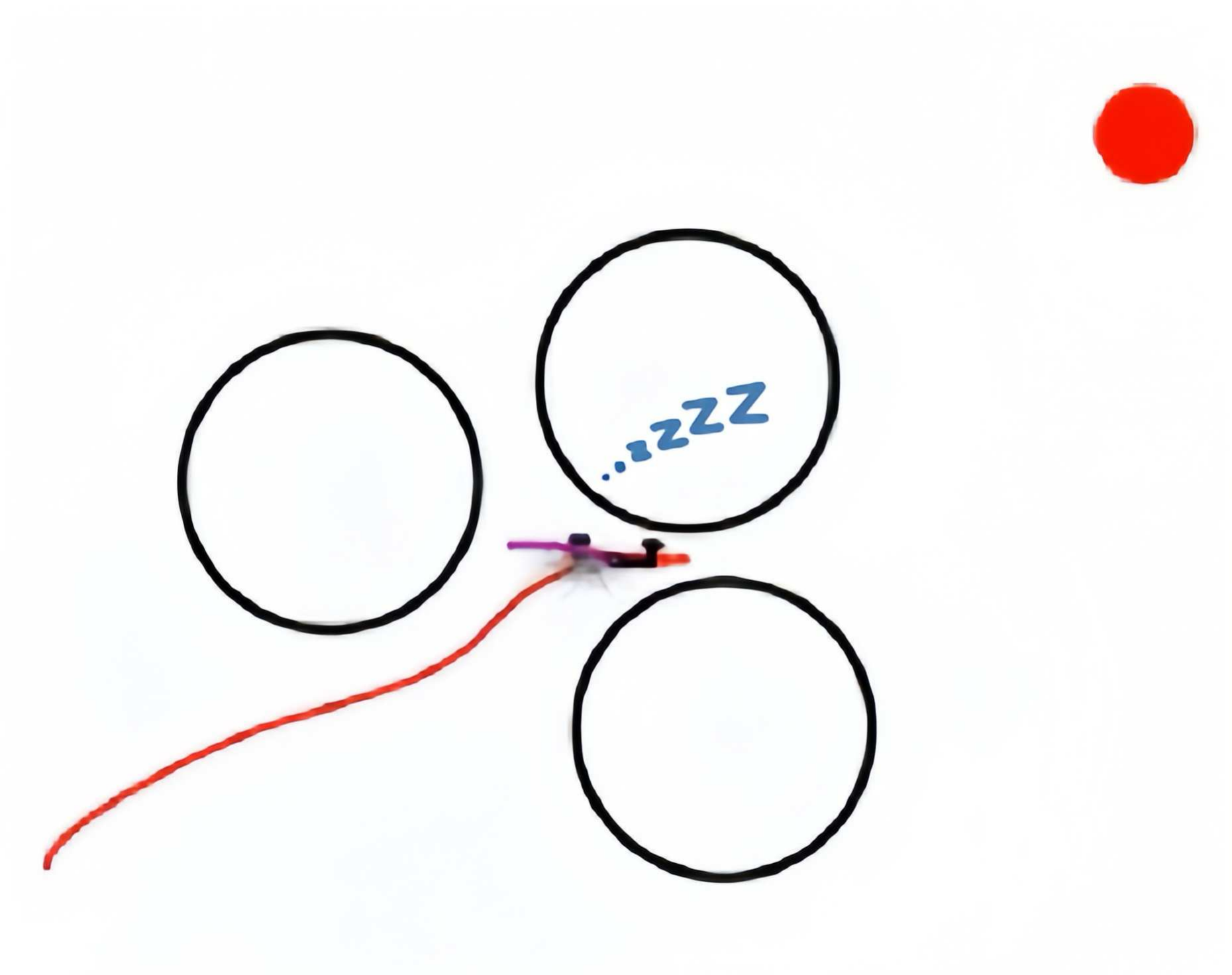} 
    \caption{Log-MPPI entrapment.}
    \label{fig:sleep}
\end{subfigure}
\vspace{0.5cm} 
\begin{subfigure}{0.235\textwidth}
    \centering
    \includegraphics[width=\textwidth]{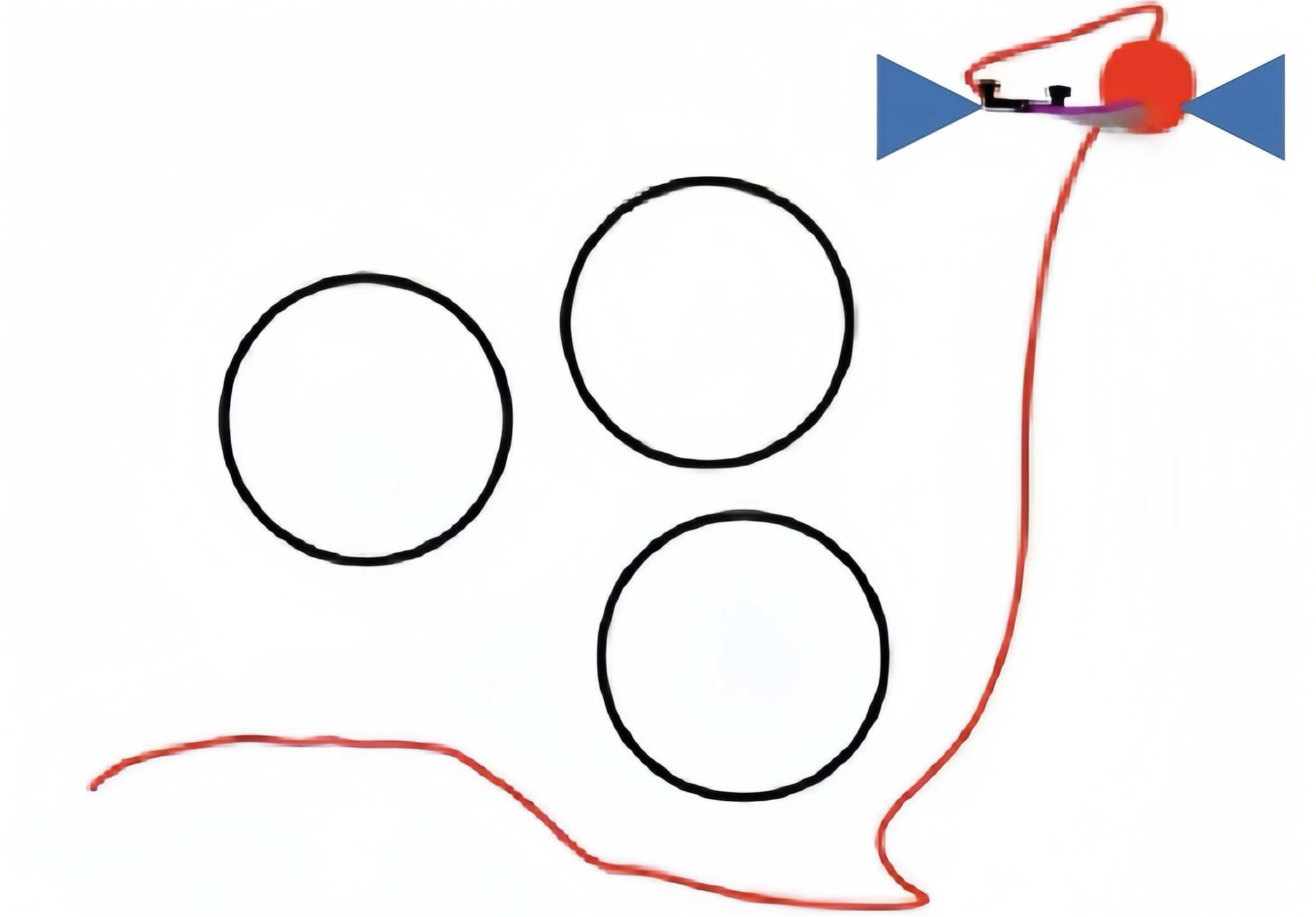} 
    \caption{Log-MPPI detour.}
    \label{fig:long}
\end{subfigure} \hfill
\begin{subfigure}{0.235\textwidth}
    \centering
    \includegraphics[width=\textwidth]{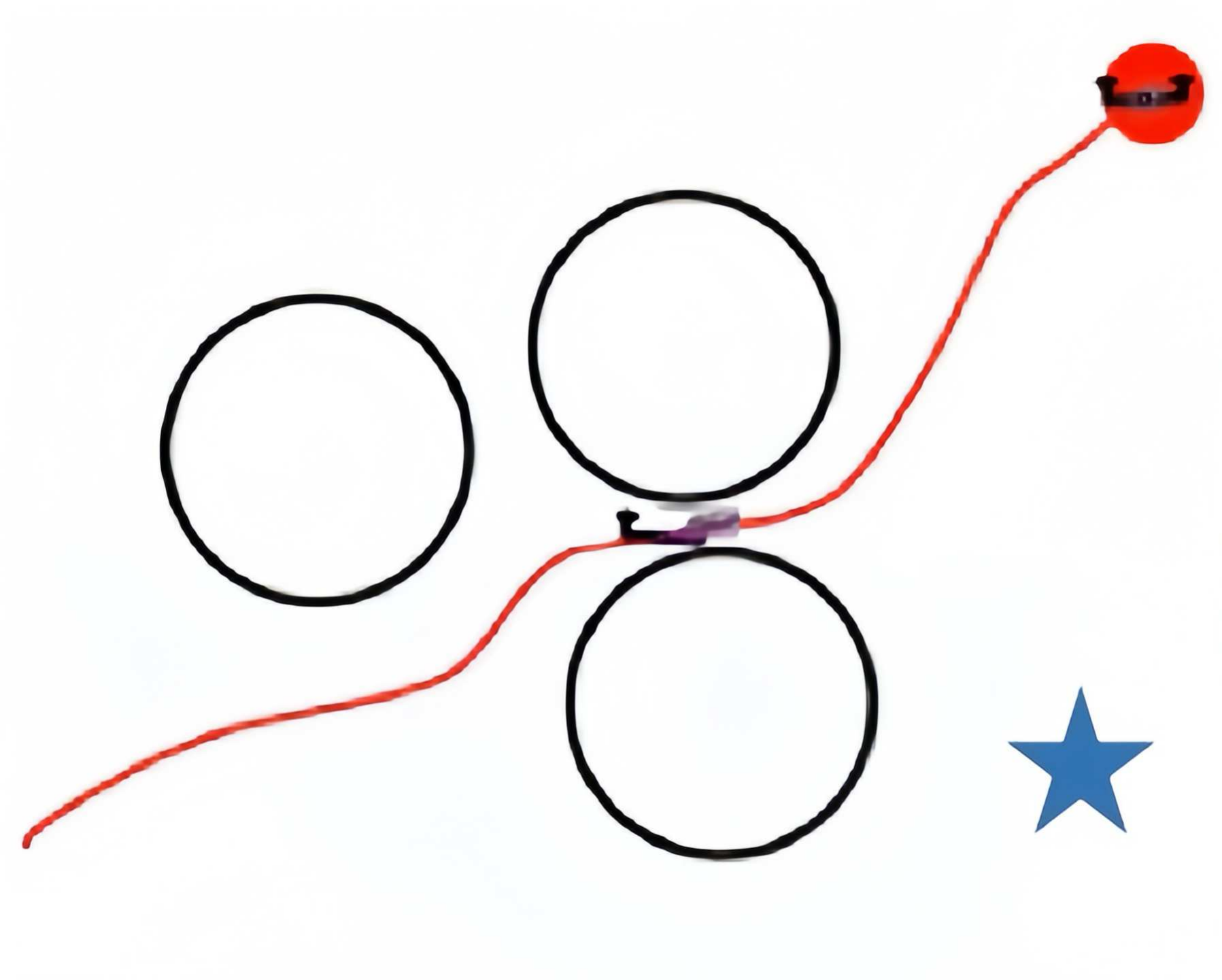} 
    \caption{DBaS-Log-MPPI success.}
    \label{fig:success}
\end{subfigure}
\caption{2D quadrotor simulation results. Black and red circles denote the obstacles and destinations, respectively. The grey, purple and red curves depict the sampled trajectories, near-optimal trajectories and executed trajectories.}
\label{fig:uav_result}
\end{figure}

The 2D quadrotor is tasked with navigating through three obstacles (each with a radius of 0.8$m$) positioned along the trajectory with narrow gaps, at a reference speed of 1$m/s$. The mission configuration and results are illustrated in Table~\ref{tab:mission_performance_comparison} and Fig.~\ref{fig:uav_result}.

Vanilla MPPI, without fine-tuning, struggles significantly in this challenging scenario, achieving a 0\% success rate with collisions occurring in all trials, as shown in Fig.~\ref{fig:coll}.
Log-MPPI achieves a modest success rate of 43\% in Mission 2. However, without the adaptive exploration proposed in our algorithm, Log-MPPI often becomes trapped in local optima (Fig.~\ref{fig:sleep}), or detour with high tracking error (Fig.~\ref{fig:long}). In contrast, our DBaS-Log-MPPI algorithm achieves a 100\% success rate, as demonstrated in Fig.~\ref{fig:success}. It exhibits the smallest tracking error (2.889$m$) and an average tracking velocity of 0.721$m/s$, which is lower than Log-MPPI due to its continuous collision risk assessment. This indicates that the DBaS implementation leads to a more conservative yet safer tracking speed, as it avoids impulse-like penalties upon constraint violations in scenarios where high speed is not demanded.

\subsection{Ground vehicle navigation simulation}

We then design simulation scenarios for the ground vehicle experiment. The vehicle is modeled as the following discrete-time nonlinear Ackermann steering model:

\begin{equation}\label{veh_dyn}
    \begin{bmatrix}
x_{k+1}\\ 
y_{k+1}\\ 
\theta_{k+1}\\
v_{k+1}
\end{bmatrix} = \begin{bmatrix}
x_{k} + v_k\cos\theta_k\Delta t\\ 
y_{k} + v_k\sin \theta_k\Delta t\\ 
\theta_{k} +  \frac{v_k\tan \phi}{L}\Delta t\\
v_k+a\Delta t
\end{bmatrix},
\end{equation}
where $\mathbf{x_k}=[x_{k}, y_{k}, \theta_{k}, v_k]$ represents the position $x$, position $y$, heading angle $\theta$ and linear velocity $v$ expressed in the world frame. The control $\mathbf{u_k}=[\phi, a]$ denotes the control signal of steering angle and acceleration with limits of $[\pm 1.013~rad, \pm 2~m/s^2]$. We propose a specific vehicle model for this purpose, where collision detection is facilitated by defining eight shape points on the vehicle's geometric model—a rectangle with a length of 4 meters and a width of 3 meters. These shape points are positioned at the corners and midpoints along each side of the rectangle. The obstacle function is defined as:
\begin{equation}
    \begin{aligned}
   h(x_i) =\| x_i - x_c^j \|_2^2 - r_c^j{^2} \geq 0, \quad \text{for } &i \in \{ 1, \dots \, 8\}, \\
   &j\in \{1, \dots \}
   \end{aligned}
\end{equation}
where $x_i$ denotes the shape points and $x_c^i$ is the center of $i$th circle obstacle with radius of $r_c$.

Sampling covariance matrix $\Sigma_u$ are set as $\begin{bmatrix} 0.075& 0 \\ 0& 2 \end{bmatrix}$, $\gamma$ are set to be 2 for all three algorithms. Coarseness factor $\mu$ is set to be 0.4 in DBaS-Log-MPPI. The real-time execution is carried out on an NVIDIA GeForce GTX 3070 laptop GPU, where all algorithms were written in Python. The prediction time horizon $N$ is set to 20 steps, the sampling time $\Delta t$ is set to be 0.02$s$ and parallel sampled trajectories number $M$ is set to 1024. The average processing time is 0.015 s, with no significant differences observed among the three algorithms.

\begin{figure}[t]
    \centering
    \includegraphics[width=\linewidth]{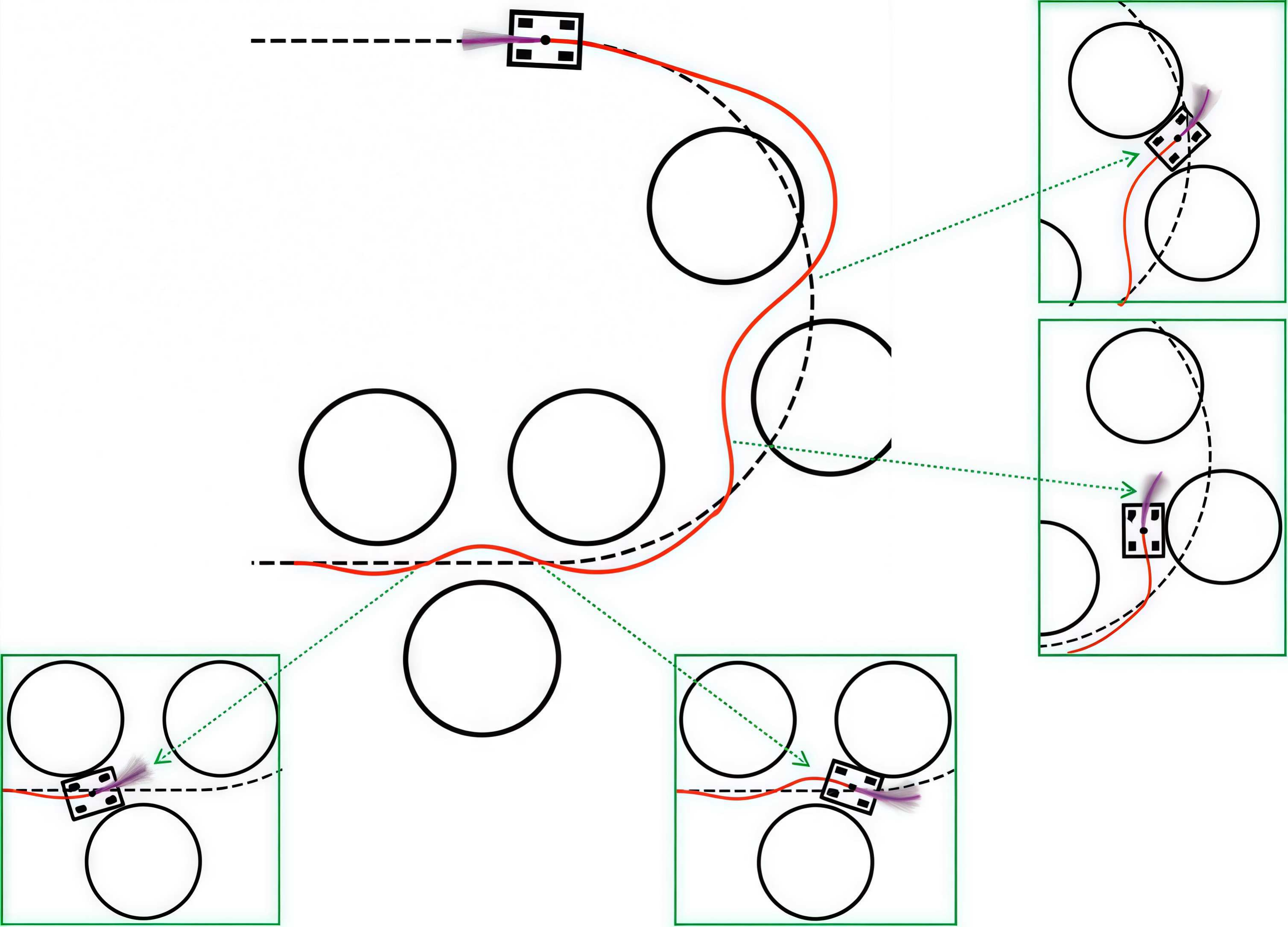} 
    \caption{Ground vehicle simulation result. Black circles denote the obstacles. The grey, purple, red and black doted curves depict the sampled trajectories, near-optimal trajectories, executed trajectory and reference trajectory, respectively.}
    \label{fig:car_result}
\end{figure}

The ground vehicle is tasked with navigating through five obstacles (each with a radius of 4.3$m$) positioned along the trajectory with narrow gaps, at reference speeds of 5$m/s$ in Mission 2 and 8$m/s$ in Mission 3, respectively. The results are summarized in Table~\ref{tab:mission_performance_comparison}. Details of obstacle avoidance in successful trials for Mission 2 of our algorithm are illustrated in Fig.~\ref{fig:car_result}.

In Mission 2, both Log-MPPI and the proposed algorithm outperform Vanilla MPPI, achieving success rates of 74\% and 100\%, respectively. Although Vanilla MPPI achieves the highest tracking speed due to its coarse search strategy, our proposed algorithm demonstrates the lowest tracking error, further emphasizing its robust performance assurance.

In Mission 3, the reference speed is more relentless at 8 m/s. Both our algorithm and Log-MPPI exhibit lower success rates compared to Mission 2, while Vanilla MPPI fails completely. On successful trials, the tracking errors for both our algorithm and Log-MPPI are reduced, as the vehicle can recover to the reference trajectory more quickly at higher speeds. However, Log-MPPI approaches obstacles at shorter distances, resulting in smaller tracking error and low average speed due to sharp turns and abrupt detours. 

\subsection{Ground vehicle real-world demonstration}

\begin{figure}[t]
\centering
\vspace{2mm}
\begin{subfigure}{0.235\textwidth}
    \centering
    \includegraphics[width=\textwidth, height=3.3cm]{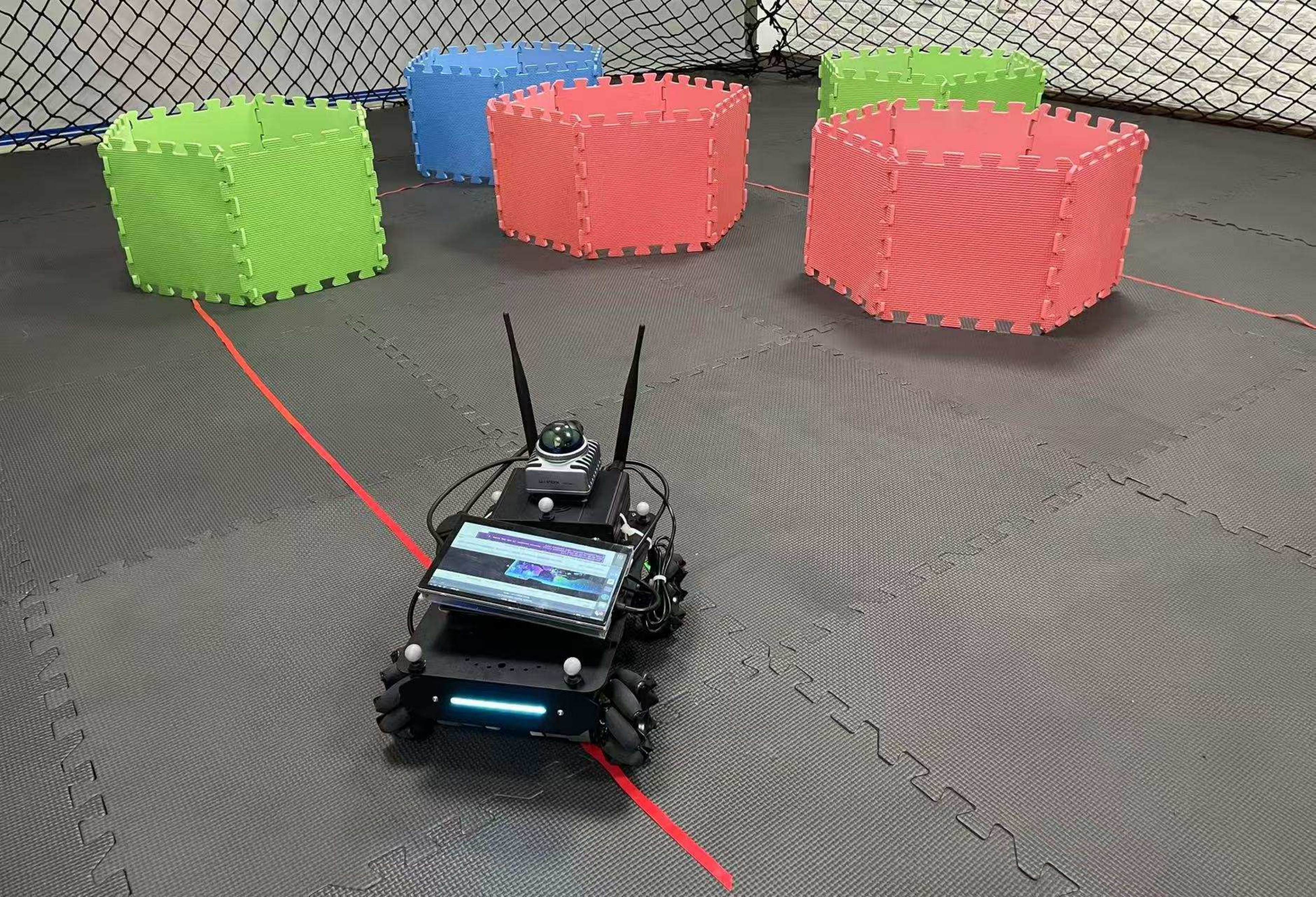} 
    \caption{Ground vehicle "Antelope".}
    \label{fig:Antelope}
\end{subfigure} \hfill
\begin{subfigure}{0.235\textwidth}
    \centering
    \includegraphics[width=\textwidth,height=3.3cm]{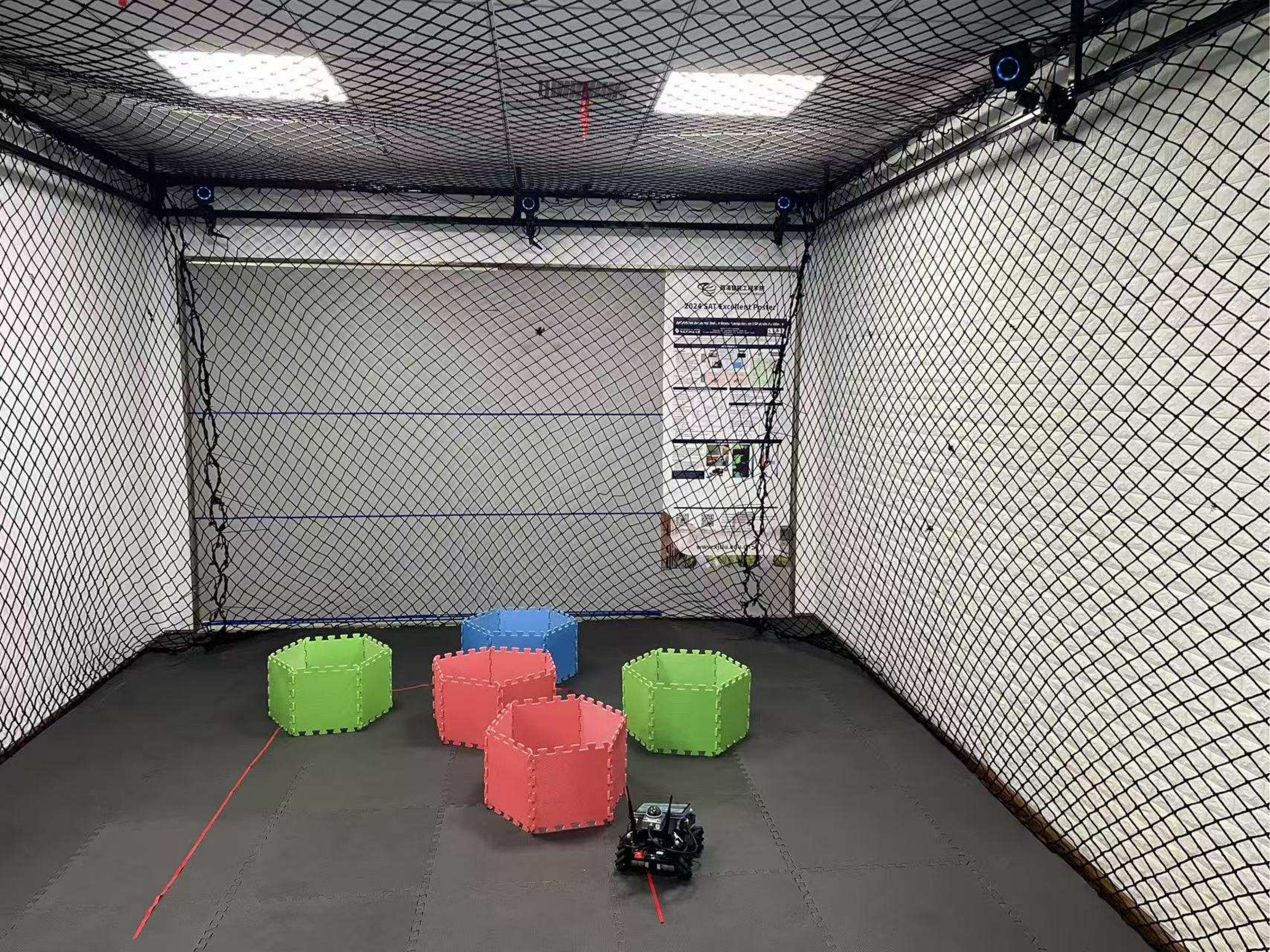} 
    \caption{Luster captures trajectory.}
    \label{fig:FZMotion}
\end{subfigure}
\vspace{0.5cm} 
\begin{subfigure}{0.235\textwidth}
    \centering
    \includegraphics[width=\textwidth,height=4cm]{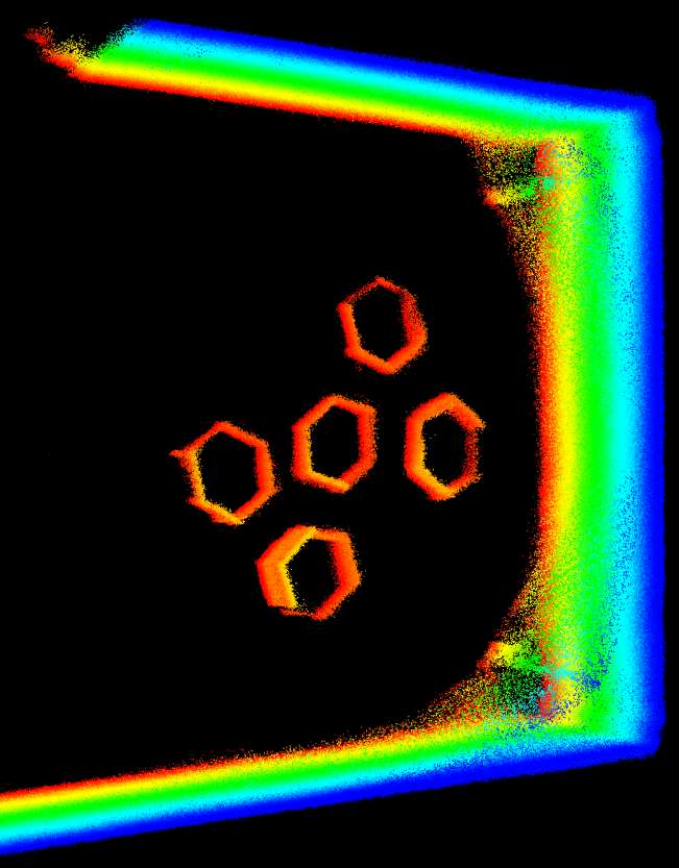} 
    \caption{Point cloud visualization.}
    \label{fig:pcd}
\end{subfigure} \hfill
\begin{subfigure}{0.235\textwidth}
    \centering
    \includegraphics[width=\textwidth,height=4cm]{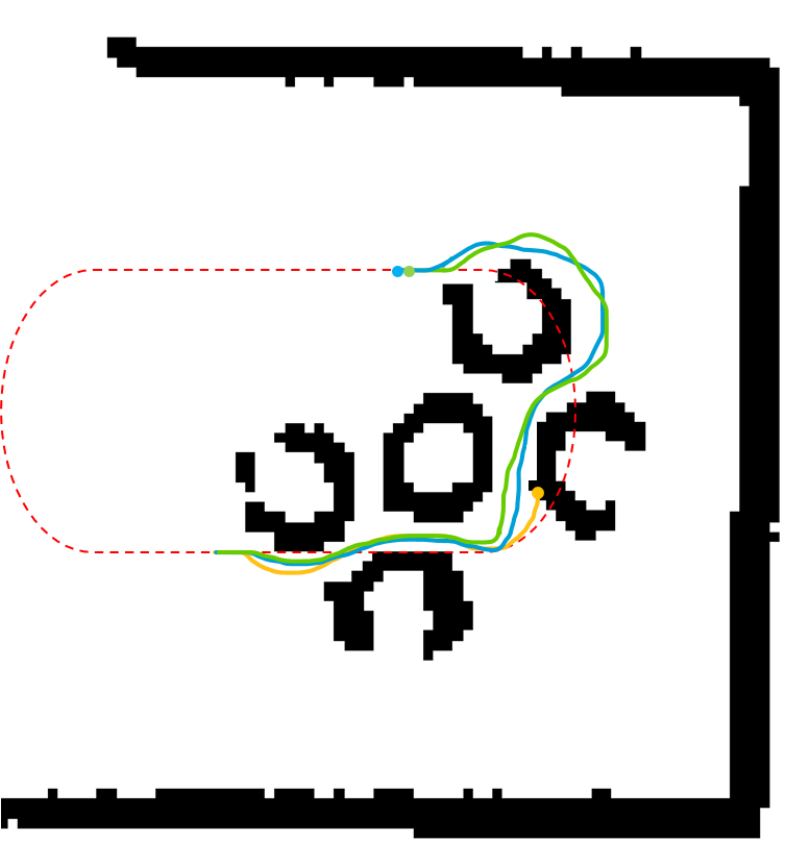} 
    \caption{Grid map with 3 trails.}
    \label{fig:gridmap}
\end{subfigure}
\vspace{-0.3cm} 
\caption{Real-world experiment details and trial comparisons for DBaS-Log-MPPI (green), Vanilla-MPPI (yellow), and Log-MPPI (blue).}
\label{fig:real result}
\vspace{-0.5cm} 
\end{figure}

Subsequently, we experimentally validated the proposed control strategies in real-world on our ground vehicle, "Antelope" (Fig.~\ref{fig:Antelope}), for achieving Mission 2 as collision-free navigation in a 2.5$m$ $\times$ 1.9$m$ indoor cluttered environment without prior environmental information.

As illustrated in Fig.~\ref{fig:FZMotion}, the ground vehicle ("Antelope"), obstacles (constructed using colorful pads), and planned trajectory (marked by red curves) are scaled down by a factor of approximately 1:16 from the simulation scenarios in Mission 2/3. Note that the "Antelope" employs Mecanum wheels; thus, we emulate an Ackermann steering system by applying differential outputs to its original kinematic model. The Luster FZMotion motion capture system (cameras mounted at a high elevation in Fig.~\ref{fig:FZMotion}) is used to obtain ground truth position of obstacles and state value of "Antelope" for post-experiment performance evaluation only. 

A LiDAR sensor (Livox Mid-360) integrated with FAST-LIO2 \cite{xu2022fast} is deployed on the "Antelope" to construct a local point cloud map, enabling obstacle pose estimation during the experiment (Fig.~\ref{fig:pcd}). The point cloud map is then converted into the 2D grid map for real-time trajectory optimization. All algorithms are deployed on an NVIDIA Jetson Orin Nano with 67 TOPS on 1024 GPU cores and a 1.7 GHz CPU frequency. The trajectories are visualized as follows: Vanilla MPPI as yellow curve, Log-MPPI as blue curve, and our algorithm as green curve. Similar to the Mission 2 results, Vanilla MPPI frequently fails in obstacle avoidance, whereas our algorithm surpasses Log-MPPI with lower tracking error (Fig.~\ref{fig:gridmap}).

\section{CONCLUSIONS}

This paper presents a novel DBaS-Log-MPPI controller that adaptively explores the action space through effective sampling within barrier states embedded in dynamics ensuring safety guarantees. The integration of the barrier state enables the proposed algorithm with continuous collision risk assessment. Furthermore, the adaptive exploration mechanism enhances sampling diversity and coverage, particularly in proximity to obstacles. Extensive validation through three simulation missions and one real-world experiment demonstrates that the proposed algorithm achieves a higher success rate, lower tracking errors, and more conservative tracking velocities compared to baseline methods.

Future work will focus on refining the controller by developing more comprehensive safety barrier formulations and extending its application to complex control scenarios, such as multi-agent planning. Additionally, we plan to implement the algorithm on advanced robotic platforms, including 3D quadrotors and quadruped robots, which present broader action spaces and challenges in sampling effective trajectories.

\addtolength{\textheight}{-1cm}   








\bibliographystyle{ieeetr}
\bibliography{yikun, yikun-planning, yikun-RL}
\end{document}